%% file: wipsnet.tex
\documentclass{article}

\usepackage{microtype}
\usepackage{graphicx}
\usepackage{subcaption}
\usepackage{booktabs}
\usepackage{hyperref}
\usepackage{multirow}

\usepackage[accepted]{icml2026}
\usepackage{amsmath}
\usepackage{amssymb}
\usepackage{mathtools}
\usepackage[capitalize,noabbrev]{cleveref}

\icmltitlerunning{WIPSNet: Deep Learning for Paediatric Wheeze Detection from Impedance Pneumography}

\begin{document}

\twocolumn[
  \icmltitle{WIPSNet: Deep Learning for Paediatric Wheeze Detection \\
    from Overnight Impedance Pneumography}

  \icmlsetsymbol{equal}{*}

  \icmlsetsymbol{equal}{*}

\begin{icmlauthorlist}
    \icmlauthor{Felix Oury}{impcomp,impbio}
    \icmlauthor{Harley Day}{impbio}
    \icmlauthor{Karina Mayoral}{nhli}
    \icmlauthor{Ville-Pekka Seppä}{icare}
    \icmlauthor{Sejal Saglani}{nhli}
    \icmlauthor{Reiko J. Tanaka}{impbio}
\end{icmlauthorlist}

\icmlaffiliation{impcomp}{Department of Computing, Imperial College London, London, United Kingdom}

\icmlaffiliation{impbio}{Department of Bioengineering, Imperial College London, London, United Kingdom}

\icmlaffiliation{nhli}{National Heart and Lung Institute and Centre for Paediatrics and Child Health, Imperial College London, London, United Kingdom; Department of Respiratory Paediatrics, Royal Brompton Hospital, London, United Kingdom}

\icmlaffiliation{icare}{Icare Finland Ltd, Vantaa, Finland}

\icmlcorrespondingauthor{Reiko Tanaka}{r.tanaka@imperial.ac.uk}

\icmlkeywords{Machine Learning, ICML}

\vskip 0.3in
  \icmlkeywords{paediatric asthma, impedance pneumography, wavelet transform, state space models, long physiological signals}

  \vskip 0.3in
]

\printAffiliationsAndNotice{}

\input{abstract}

\input{intro}
\input{methods}
\input{results}

\input{discussion}

\bibliography{wipsnet}
\bibliographystyle{icml2026}

\newpage
\appendix
\onecolumn
\input{appendix}

\end{document}

%% file: Abstract.tex
\setcounter{footnote}{1}
\begin{abstract}
Overnight impedance pneumography (IP) is used to monitor paediatric respiratory health. Its current clinical readout, the Expiratory Variability Index (EVI), compresses each IP recording into a single scalar and achieves an AUC of 0.633 for night-level wheeze classification. We introduce Wheeze Impedance Pneumography Scalogram Network (\textbf{WIPSNet}\footnote{\scriptsize Code: \url{https://github.com/felix-oury/WIPSNet}}), a 3D ResNet operating on stacked continuous wavelet transform scalograms of overnight IP signals. On a 15-patient cohort (60 nights, 281 hours), WIPSNet achieves an AUC of $0.783 \pm 0.026$, outperforming EVI, a state-space model (Mamba), and two modern sleep-staging architectures. Performance peaks at a volumetric depth corresponding to 32 minutes of temporal context, suggesting that multi-scale temporal aggregation is important for modelling nocturnal respiratory dynamics. Overall, these results indicate that structured time--frequency representations combined with 3D convolutional architectures provide an effective approach for learning from long, irregular physiological time series.
\end{abstract}

%% file: Intro.tex
\section{Introduction}
\label{sec:intro}
Asthma is the most common chronic lung disease in childhood. Later asthma risk is associated with acute wheezing episodes in preschool-age children~\citep{stern2020,saglani2007}. These wheezing episodes result from airway narrowing and bronchial obstruction, but are often difficult to recognise at home, delaying treatment and increasing morbidity. This motivates the need for remote paediatric wheeze monitoring tools~\citep{gina2025,makrinioti2024}.

AI-powered digital stethoscopes can detect wheeze in controlled settings~\citep{cheng2022,habukawa2020}, but require active cooperation from the child, making them unsuitable for continuous overnight monitoring. Impedance pneumography (IP) is a viable alternative that measures thoracic electrical impedance~\citep{seppa2016}. The Ventica\textsuperscript{\textregistered} system is a commercially deployed device that summarises an overnight IP recording into a single scalar, Expiratory Variability Index (EVI). However, EVI's clinical utility is limited by high inter-individual variability~\citep{seppa2020evi,burman2021,seppa2020reduced}. A recent systematic review~\citep{children13020193} concludes that the EVI is better suited for longitudinal monitoring of airway function than for wheeze diagnosis. This motivates the development of novel methods for wheeze diagnosis that exploit richer information contained in the IP waveform. In addition, overnight IP recordings are irregular due to motion artefacts and sensor displacement, resulting in variable-length and partially corrupted time series.

Deep learning can exploit the full IP waveform, but the extreme length ($\sim 10^7$ samples) and irregularity of overnight IP recordings challenge conventional architectures. Recurrent neural networks suffer from vanishing gradients ~\citep{pascanu2013}, transformers scale quadratically~\citep{vaswani2017}, and state-space models such as Mamba~\citep{gu2024mamba} have limited long-context capabilities ~\citep{waleffe2024empiricalstudymambabasedlanguage,benkish2025decimambaexploringlengthextrapolation,ye2025longmambaenhancingmambaslong}. In contrast, long physiological signals favour convolutional backbones operating on time-frequency inputs~\citep{ijerph19106322, rs12101685, biomedinformatics5010007}. These approaches can be viewed as learning structured representations that capture local spectral patterns and their evolution over longer time scales. Additionally, modern sleep-staging architectures, such as AttnSleep~\citep{9417097} and SleePyCo~\citep{Lee_2024}, provide strong baselines for our task as they are designed for similar long, continuous physiological signals recorded during sleep.

We introduce \textbf{WIPSNet}, a deep learning pipeline for paediatric nocturnal wheeze classification from overnight IP recordings. Building on a 3D ResNet over stacked continuous wavelet (CWT) scalograms, WIPSNet exploits the full IP waveform through a modular design that accommodates variable-length and partially corrupted recordings. We benchmark against EVI, AttnSleep, SleePyCo, and Mamba, achieving an AUC improvement over EVI ($+0.15$) and Mamba ($+0.094$). Our results suggest that explicitly structured time--frequency representations combined with convolutional inductive biases are particularly effective for modelling long, oscillatory physiological signals, and suggest that SSMs may struggle to fully exploit multi-scale structure in this setting.

%% file: Methods.tex
\section{Methods}
\label{sec:methods}

\subsection{Data and Labels}
Overnight IP recordings were collected using the Ventica\textsuperscript{\textregistered} system from 15 preschool children (mean age $3.3 \pm 1.3$\,years; 60 nights; 281.3 hours; 250\,Hz) in two London (UK) hospitals.  Each night was labelled either wheezy or non-wheezy depending on whether WheezeScan\textsuperscript{\textregistered}, a clinically validated acoustic device~\citep{habukawa2020,omron2025}, detected wheeze or not, prior to sleep or upon waking. The cohort contains 21 wheezy and 39 non-wheezy nights.

\subsection{WIPSNet}
\begin{figure*}
    \centering
    \small
\includegraphics[width=0.8\textwidth]{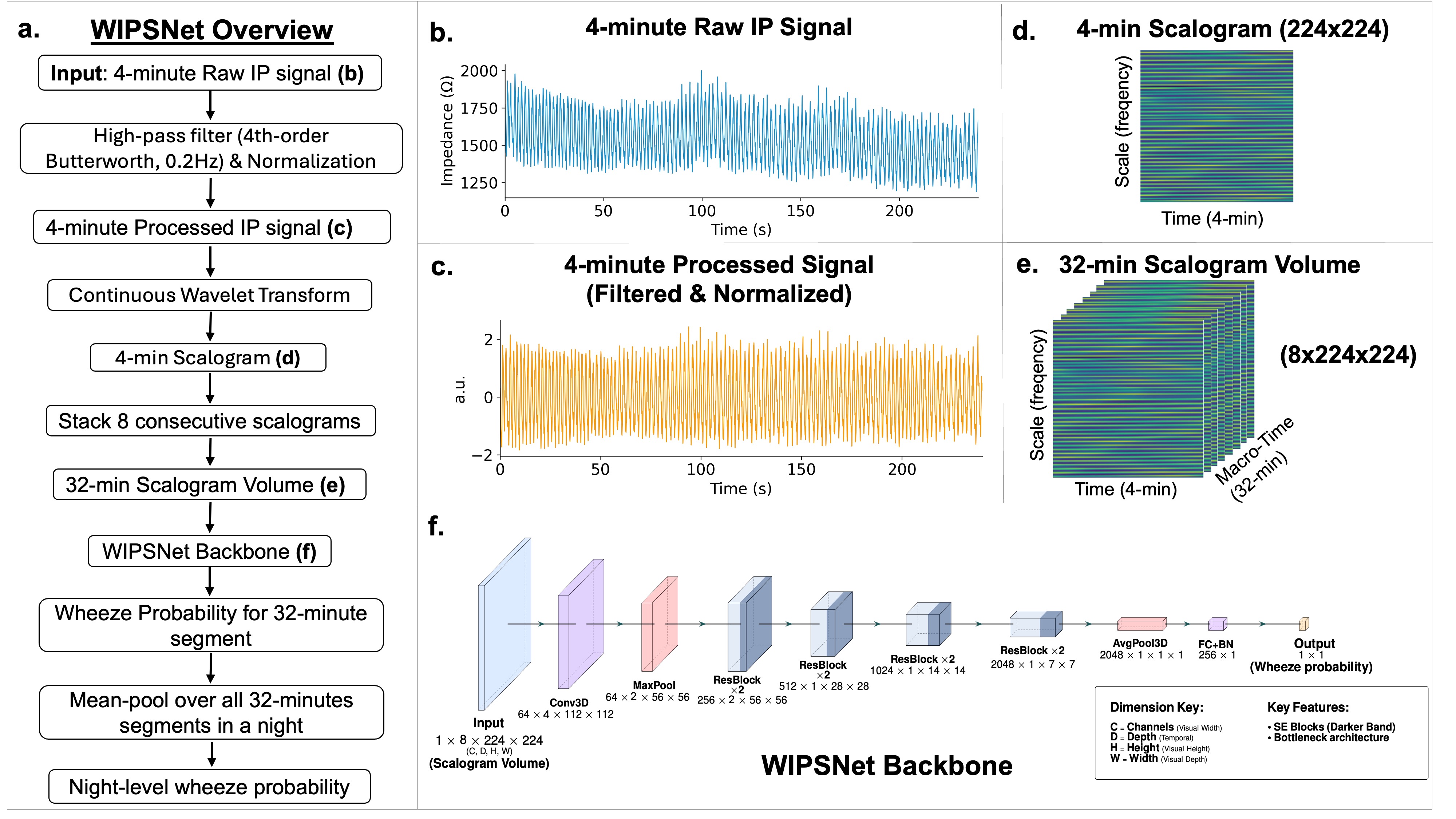}
    \caption{WIPSNet framework and architecture. (a) The IP signal is filtered, transformed into scalograms, grouped into 3D volumes, and passed through a 3D CNN; volume-level probabilities are aggregated into a night-level prediction. (b--d) Raw signal, processed signal, and 4-minute scalogram. (e) Volumetric input construction: $D$ consecutive scalograms stacked along the depth axis. (f) The bottleneck 3D ResNet backbone.}
    \label{fig:wipsnet_arch}
\end{figure*}

WIPSNet (\cref{fig:wipsnet_arch}) takes a 32-minute segment of overnight IP signal at 250\,Hz as input and outputs a wheeze probability for that segment. The pipeline processes raw IP signal through high-pass filters, transforms it into consecutive time-frequency scalograms, stacks them into 3D volumes, and processes them with a 3D CNN to predict wheeze. Night-level predictions are obtained by mean-pooling segment-level outputs, enabling variable-length recordings.

Each overnight recording is high-pass filtered (4th-order Butterworth, 0.2\,Hz) to remove baseline drift, then split into non-overlapping 4-minute windows (60{,}000 samples) (\cref{fig:wipsnet_arch}b--c).
Each window is Z-score normalised and transformed via the continuous wavelet transform (CWT) into a 224$\times$224 scalogram using a complex Morlet wavelet (\cref{fig:wipsnet_arch}d). Eight consecutive scalograms are then stacked to form a 3D volume 
representing $8\times 4=32$ minutes of continuous IP signal (\cref{fig:wipsnet_arch}e). The volume depth, $D{=}8$, was selected via hyperparameter optimisation.

This design explicitly encodes a two-timescale structure in the data: short-term spectral content within scalograms and longer-term evolution across stacked windows. Within each scalogram, the 2D time-frequency representation captures breath-by-breath spectral content over a 4-minute window. The stacked scalograms encode how the spectral content evolves over 32 minutes, capturing the onset, persistence, and resolution of respiratory events. A 3D convolution over the 32-minute scalogram volume integrates these two time-scales into a unified representation of respiratory dynamics.

Each scalogram volume is processed by a 3D ResNet (Bottleneck3D, $[2,2,2,2]$ layers, expansion~4)~\citep{he2016resnet} (\cref{fig:wipsnet_arch}f). Global average pooling is followed by a fully connected layer with batch normalisation (256 units), ReLU and dropout, and a final linear layer that outputs a wheeze logit per volume (\cref{fig:wipsnet_arch}f). All hyperparameters were tuned via Optuna~\citep{akiba2019optuna} (50 trials).

\subsection{Baselines for benchmarking}

{\bf State-space models:}
Mamba~\citep{gu2024mamba} was trained directly on raw 1D IP under the same Optuna HPO budget used for WIPSNet. To probe context scaling, the SSM analogue of WIPSNet's volumetric depth sweep, we trained Mamba on raw windows of $\{4, 8, 16, 32\}$ minutes. The 32-minute window matches WIPSNet's temporal field of view and yields the best Mamba result, which we report in \cref{tab:main}. To isolate whether the CWT representation alone suffices, \emph{Mamba-Scalogram} feeds Mamba WIPSNet's scalogram volumes via ViT-style patch tokenisation (frames resized to $192{\times}192$, $16{\times}16$ patches, 1{,}152 tokens per volume).

{\bf Modern single-channel sleep-staging architectures:}
AttnSleep~\citep{9417097} is a multi-resolution 1D CNN with adaptive feature recalibration and causal self-attention.  SleePyCo~\citep{Lee_2024} is a multi-scale feature pyramid and Transformer classifier. Both operate on raw single-channel 1D physiological signal at 100\,Hz with epoch-based aggregation, extract multi-scale temporal features without EEG-specific priors (no band filters), and target the same regime of ultra-long overnight recordings with subject-wise cross-validation. We use published default configurations, replacing only the final layer with a binary classification head. Epoch probabilities are mean-pooled into a night-level prediction. Each 4-minute IP segment is resampled from 250\,Hz to 100\,Hz and split into eight 30-second sub-epochs. This matches the original architectures' expected input dimensions, ensuring that their default convolutional receptive fields accurately capture the intended physical time scales.

{\bf Non-deep-learning baselines:}
We include EVI as the standard clinical baseline and two classical ML baselines built on an identical gradient-boosted-tree classifier (XGBoost), so the only variable between them is the input representation. The first, \emph{XGBoost (spectral)}, uses 44 hand-crafted summaries of the CWT scalogram: per-band mean energy and temporal variance over 20 contiguous frequency bands, the high/low band energy ratio, overall mean energy, and the spectral centroid and bandwidth. The second, \emph{catch22}, uses the 22 canonical, domain-agnostic time-series features of \citet{lubba2019catch22canonicaltimeseriescharacteristics} computed directly on the raw 1D impedance signal. Both featurise at the 4-min window level, train with positive-class re-weighting under patient-level LOOCV, and aggregate to night-level predictions by averaging window probabilities, matching the WIPSNet protocol. Both discard fine local time--frequency structure, which is the gap the convolutional model is hypothesised to exploit.

{\bf Representation ablations:}
We performed representation and context ablations to validate WIPSNet's specific design choices. We replaced the CWT front-end with a short-time Fourier transform, and wavelet scattering, keeping the 3D ResNet backbone identical. We examined the volumetric depth $D \in \{1, 2, 4, 8, 12, 16, 24, 32\}$ for the CWT variant to quantify the impact of temporal context (ranging from 4 to 128 minutes of continuous signal).

\subsection{Evaluation}
All models and baselines were evaluated using the patient-level leave-one-out cross-validation (LOOCV, 15 folds) across 5 random seeds, using night-level aggregation. While the cohort is small, LOOCV maximises training data and results are averaged across 5 seeds to ensure robustness. We report two threshold-independent metrics: Area Under the Receiver Operating Characteristic (AUC) and Area Under the Precision-Recall Curve (AUPRC). AUC captures global class separability, while AUPRC focuses on performance on the positive class, which is clinically meaningful since missing a wheezing episode is a missed opportunity to treat a patient. 
A random classifier achieves an AUPRC of 0.35 for our cohort prevalence of 35\% positive nights (21/60).

%% file: Results.tex
\section{Results}
\label{sec:results}

WIPSNet achieves an AUC of $0.783 \pm 0.026$ and an AUPRC of $0.634 \pm 0.047$ (\cref{tab:main}). The improvement of WIPSNet over EVI ($+0.15$ AUC, $+0.201$ AUPRC) suggests that the overnight IP recording contains discriminative information that the EVI discards. EVI was not computable on 3/60 nights because the sleep duration did not meet the algorithm's minimum signal-length requirement. Handcrafted CWT spectral features (XGBoost, $0.553$ AUC) lack the temporal context to be competitive, yet generic canonical time-series statistics (catch22, $0.704$ AUC) form a strong baseline that surpasses both Mamba and the Transformer-based SleePyCo. WIPSNet and AttnSleep lead the table at $0.783$ and $0.764$ AUC, while Mamba reaches $0.689$ and SleePyCo drops to $0.577$.

WIPSNet's lead over the strongest baseline, AttnSleep, is modest on AUC ($+0.019$) and somewhat larger on AUPRC ($+0.031$). The wider AUPRC margin is consistent with better handling of the imbalanced minority wheeze-positive class, though the two models remain close. The high seed variance of SleePyCo ($\pm0.152$ AUC) and Mamba-Scalogram ($\pm0.138$) relative to WIPSNet ($\pm0.026$) further indicates that the convolutional front-end yields more stable predictions across seeds.

WIPSNet processes an overnight recording in $\sim0.34$\,s on a consumer GPU (NVIDIA RTX 4070), providing more than enough throughput for the cloud-based retrospective analysis of longitudinal wearable data.
\begin{table}[t]
\centering
\scriptsize
\caption{Night-level wheeze classification results. Mean $\pm$ std over 5 seeds, 15-fold patient-level LOOCV. AUPRC random baseline at our class prevalence is 0.35. Mamba and Mamba-Scalogram both use a 32-min window. $^\dagger$Deterministic; 57/60 valid nights.}
\label{tab:main}
\setlength{\tabcolsep}{4pt}
\begin{tabular}{llcc}
\toprule
Model & Input \& backbone & AUC & AUPRC \\
\midrule
\textbf{WIPSNet} & CWT + 3D CNN & $\mathbf{.783{\scriptstyle\pm.026}}$ & $\mathbf{.634{\scriptstyle\pm.047}}$ \\
AttnSleep & Raw + MRCNN-attn & $.764{\scriptstyle\pm.030}$ & $.603{\scriptstyle\pm.048}$ \\
catch22+XGBoost & 22 canonical TS features & $.704{\scriptstyle\pm.002}$ & $.481{\scriptstyle\pm.002}$ \\
Mamba & Raw + SSM & $.689{\scriptstyle\pm.047}$ & $.509{\scriptstyle\pm.060}$ \\
EVI$^\dagger$ & Clinical scalar & $.633$ & $.433$ \\
SleePyCo & Raw + Pyramid-Transformer & $.577{\scriptstyle\pm.152}$ & $.444{\scriptstyle\pm.101}$ \\
Mamba-Scalogram & CWT patches + SSM & $.569{\scriptstyle\pm.138}$ & $.441{\scriptstyle\pm.104}$ \\
XGBoost (spectral) & Handcrafted (44 feat) & $.553{\scriptstyle\pm.010}$ & $.402{\scriptstyle\pm.006}$ \\
\bottomrule
\end{tabular}
\end{table}

\subsection{Volumetric Depth as a Driver of Performance}
\label{sec:depth}
\begin{figure}[t]
\centering
\includegraphics[width=0.8\columnwidth]{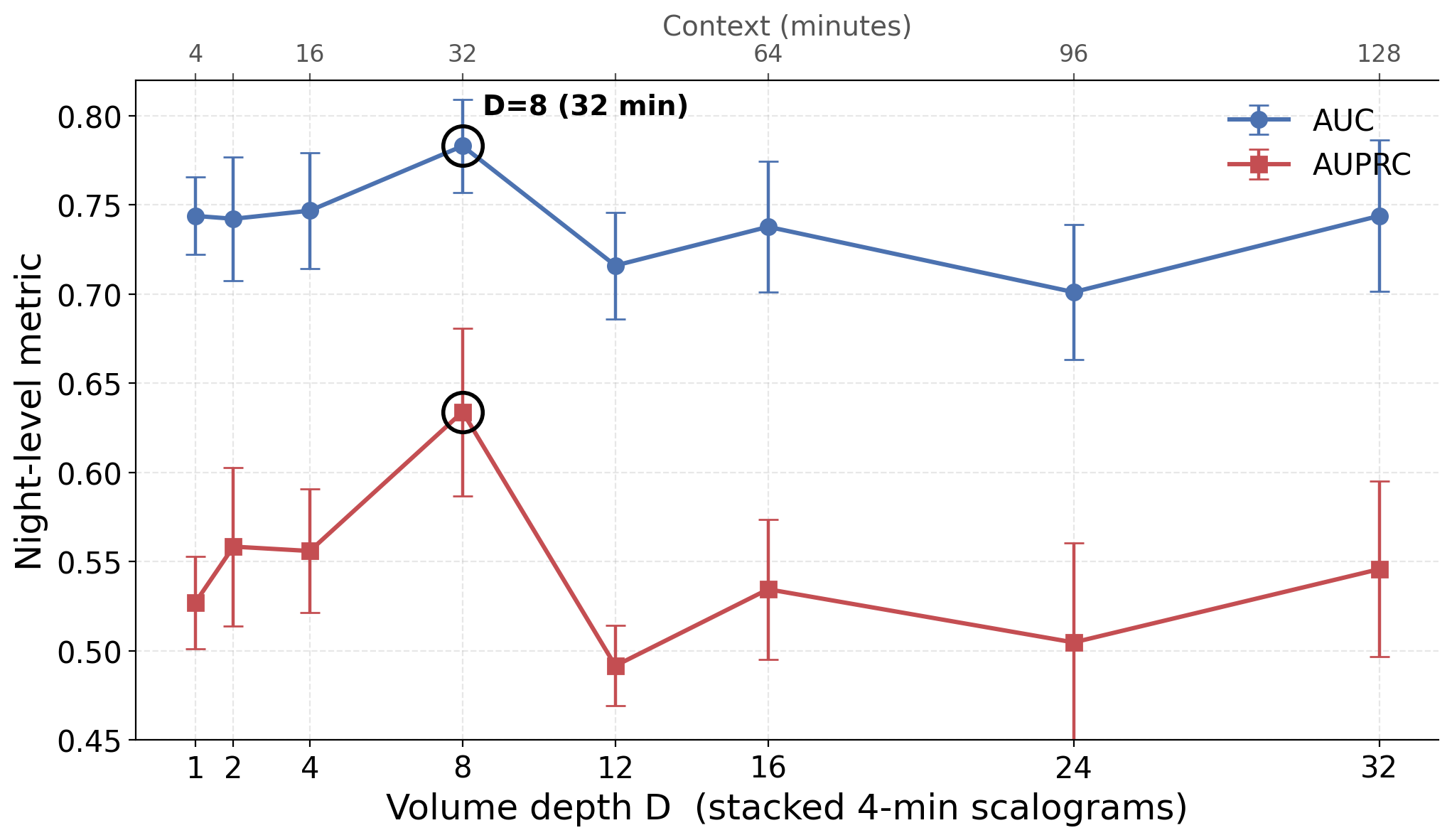}
\caption{WIPSNet's depth-scaling behaviour. Both AUC and AUPRC peak at $D{=}8$ (32\,min of temporal context) and decline at longer windows, indicating an optimal macro-timescale rather than monotone improvement. Error bars: standard deviation over 5 seeds. Top axis: temporal context in minutes.}
\label{fig:depth}
\end{figure}

Volumetric depth $D$ controls the 3D CNN's temporal context, as $D$ stacked 4-minute scalograms span $4D$ minutes. Sweeping $D \in \{1, 2, 4, 8, 12, 16, 24, 32\}$ (\cref{fig:depth}), we find a clear optimum at $D{=}8$ (32\,min). Below this, short windows of 4--16\,min are roughly flat at AUC $\approx 0.74$ and AUPRC $\approx 0.55$, indicating insufficient temporal context. At $D{=}8$ both metrics peak together, AUC $0.783 \pm 0.026$ and AUPRC $0.634 \pm 0.047$, a gain of $+0.04$ AUC and $+0.08$ AUPRC over the short-window regime. Beyond 32\,minutes performance degrades and does not recover out to 128\,minutes. This suggests the discriminative respiratory activity spans only a few tens of minutes: larger windows increasingly mix wheeze and non-wheeze breathing, blurring the signal.

This profile supports WIPSNet’s two-timescale design: 2D scalograms
capture breath-level spectra, while volumetric depth tracks their evolution over tens of minutes to mirror nocturnal oscillatory events. Fixed-sequence baselines (AttnSleep and SleePyCo) lack an explicit mechanism for such hierarchical temporal aggregation, which may limit their ability to model
extended physiological dynamics.

\subsection{Ablating Context and Representation in Mamba}
\label{sec:mamba}
We conduct two controlled experiments (Appendix~\ref{app:full}) to ask whether Mamba's shortfall relative to the convolutional model stems from limited context or from input representation. First, we sweep the raw-signal window over $\{4, 8, 16, 32\}$ minutes, the SSM analogue of WIPSNet's depth sweep. Mamba is flat and non-monotone within seed variance across the sweep (AUC $0.61$--$0.69$; Table~\ref{tab:full}), and stays $0.07$--$0.13$ AUC below the matched-context WIPSNet variant at every window. The gap is thus present already at short context rather than emerging only at long context. We attribute this to the fixed-size hidden state interacting with the high sampling rate of IP: even after downsampling, a 32-minute window spans tens of thousands of timesteps that Mamba must compress into constant memory, placing it in the regime where SSM information retention is known to degrade~\citep{waleffe2024empiricalstudymambabasedlanguage,benkish2025decimambaexploringlengthextrapolation,ye2025longmambaenhancingmambaslong}.
Second, \emph{Mamba-Scalogram} feeds Mamba the CWT volumes via patch tokenisation but reaches only AUC $0.569 \pm 0.138$ with high seed variance. Patch tokenisation linearises each 2D time--frequency scalogram into a sequence, discarding the local structure that convolution exploits. Together these results show that neither longer context nor the richer time--frequency representation alone closes the gap to the convolutional backbone.

%% file: Discussion.tex
\section{Discussion}
\label{sec:discussion}

\subsection{Clinical Impact}
WIPSNet is the first deep learning pipeline, to the best of our knowledge, for paediatric nocturnal wheeze classification from overnight IP recording. Its AUC improvement over EVI, a scalar metric with poor individual-level utility~\citep{children13020193}, is large enough ($+0.15$) to potentially make WIPSNet a clinical decision-support tool. The sliding aggregation of 4-minute volumes makes WIPSNet independent of recording length, a practically important property since sleep duration varies night to night. Inference takes $\sim0.34$\,s per night recording, making WIPSNet feasible for real-world deployment. Reliable nocturnal wheeze detection, as an objective marker of airway obstruction, could enable earlier recognition of exacerbations, inform treatment decisions from longitudinal data rather than subjective parental report, and contribute to better management of acute wheeze exacerbations in early childhood.

\subsection{ML Impact}
WIPSNet ($0.783$ AUC) and AttnSleep ($0.764$) lead the benchmark, with the raw-signal SSM (best Mamba, $0.689$) and the Transformer-based SleePyCo ($0.577$) trailing despite comparable tuning effort. Together with the ablations in~\cref{sec:mamba}, this is consistent with documented challenges for state-space models on long, high-frequency sequences~\citep{waleffe2024empiricalstudymambabasedlanguage,benkish2025decimambaexploringlengthextrapolation,ye2025longmambaenhancingmambaslong}. Mamba's accuracy is largely insensitive to context within the tested range and never closes the gap to the convolutional model, while routing the CWT representation through patch tokens (\emph{Mamba-Scalogram}) degrades rather than recovers performance.
These findings indicate that convolutional architectures remain a strong baseline for long, irregular, oscillatory physiological signals, as they preserve local time--frequency patterns while aggregating them hierarchically across time. In contrast, a fixed-state SSM operating on the raw high-frequency waveform must compress the entire sequence into constant memory and appears to lose the local structure that discriminates wheeze. WIPSNet's gains over AttnSleep are smaller in AUC but more pronounced in AUPRC, and its performance suggests that explicit, tunable hierarchical aggregation is beneficial when the relevant dynamics occupy a bounded macro-timescale.

\subsection{Limitations and Future Work}
While the dataset comprises over 280 hours of high-frequency recording ($\sim$250 million samples), it remains small at the cohort level (15 children, 60 nights). External validation in larger and more diverse populations is required. Furthermore, labels derived from WheezeScan\textsuperscript{\textregistered} at sleep boundaries may miss transient intra-night wheezing and are subject to measurement noise from the reference device. Addressing these limitations with larger cohorts and finer-grained labels would enable modelling of intra-night dynamics, including prediction of wheeze onset. Integrating these predictions with clinical data such as medication intake could further support personalised treatment strategies. Extending the framework to longitudinal multi-night modelling, together with improved wearable acquisition, would advance objective respiratory monitoring in home settings.

\section*{Impact Statement}

This work develops a deep learning approach for nocturnal wheeze detection from overnight impedance pneumography, with potential to support earlier identification of respiratory exacerbations and improve home-based monitoring in young children. Such tools could contribute to more objective and timely management of wheezing disorders. However, the model is trained on a small cohort and may not generalise without further validation. Misclassifications could lead to missed events or unnecessary concern, so the system should be used as decision support rather than a replacement for clinical judgment. Continuous physiological monitoring also raises data privacy considerations, particularly in paediatric settings. Further validation, fairness assessment, and careful integration into clinical workflows are required before deployment.

\section*{Acknowledgment}
This project was supported by UKRI EPSRC Grant Ref: EP/W002280/1, Asthma and Lung UK, UKRI AI Centre for Doctoral Training in Digital Healthcare grant number EP/Y030974/1 and by the NIHR Imperial Biomedical Research Centre (BRC). We are grateful to TIMIK for providing the Ventica systems for this study.

%% file: Appendix.tex
\section{Hyperparameter Search Spaces and Selected Values}
\label{app:hpo}

WIPSNet and Mamba were both tuned via Optuna (50 trials each).

\begin{table}[h]
\centering
\caption{WIPSNet hyperparameter search space and selected values.}
\label{tab:hpo_wipsnet}
\setlength{\tabcolsep}{6pt}
\begin{tabular}{lll}
\toprule
\textbf{Parameter} & \textbf{Search Space} & \textbf{Selected Value} \\
\midrule
Scalogram window length & 2--16\,min & 60{,}000 (4\,min) \\
Volume depth $D$ & 1--10 scalograms & 8 \\
Stride (scalogram) & 1--$D$ scalograms & 8 ($=D$, non-overlapping) \\
Backbone & ResNet / EfficientNet & ResNet \\
Batch size & \{8, 16, 32, 64\} & 32 \\
Learning rate & $10^{-5}$--$10^{-2}$ & $3.00 \times 10^{-4}$ \\
Weight decay & $10^{-6}$--$10^{-3}$ & $1.00 \times 10^{-3}$ \\
Dropout & 0--0.5 & 0.464 \\
SE attention & \{True, False\} & False \\
Mixup $\alpha$ & 0--0.5 & 0.354 \\
LR scheduler & \{True, False\} & True \\
\bottomrule
\end{tabular}
\end{table}
\begin{table}[h]
\centering
\caption{Mamba SSM hyperparameter search space and selected values. The context window $W$ (number of consecutive 4-min segments) was swept separately, analogous to the WIPSNet volume depth $D$, while the architecture was tuned by Optuna.}
\label{tab:hpo_mamba}
\setlength{\tabcolsep}{6pt}
\begin{tabular}{lll}
\toprule
\textbf{Parameter} & \textbf{Search Space} & \textbf{Selected Value} \\
\midrule
Context window $W$ (4-min segments) & \{1, 2, 4, 8\} sweep & 8 ($=32$\,min) \\
Stride (segments) & $=W$ (non-overlapping) & 8 \\
Per-segment length & --- & 5,000 samples (4\,min)\footnotemark \\
Sequence length & $W \times$ seg.\ length & 40,000 samples (32\,min) \\
$d_\text{model}$ & 16--192 & 32 \\
$d_\text{state}$ & 8--48 & 32 \\
$d_\text{conv}$ & 2--4 & 3 \\
Expand factor & 1--4 & 3 \\
Num.\ layers & 2--8 & 3 \\
Dropout & 0--0.5 & 0.374 \\
Batch size & \{8, 16, 32, 64\} & 16 \\
Learning rate & $10^{-5}$--$10^{-2}$ & $5.35 \times 10^{-4}$ \\
Weight decay & $10^{-6}$--$10^{-3}$ & $2.54 \times 10^{-4}$ \\
Scheduler & \{True, False\} & False \\
\bottomrule
\end{tabular}
\end{table}
\footnotetext{The raw 250\,Hz IP signal is downsampled $12\times$ to $\approx$20.8\,Hz before Mamba, so a 4-min segment is 5{,}000 samples and a 32-min window is $8\times5{,}000=40{,}000$ samples. This keeps the SSM sequence length tractable; on the full-rate 480{,}000-sample window Mamba collapses to near-chance.}

\pagebreak

\section{Full Results Table}

\label{app:full}
\begin{table}[h]
\centering
\caption{All models and ablations, sorted by AUC. Mean $\pm$ std over 5 seeds for DL models. AUC is the primary ranking metric; AUPRC reflects performance on the minority (wheeze) class under class imbalance. Depth-sweep rows ($D{\neq}8$) and front-end rows share WIPSNet's backbone; the headline model is WIPSNet-CWT ($D{=}8$).}
\label{tab:full}
\setlength{\tabcolsep}{4pt}
\begin{tabular}{llcc}
\toprule
Model & Input \& backbone & AUC & AUPRC \\
\midrule
\textbf{WIPSNet-CWT ($D{=}8$)} & CWT + 3D CNN                   & $\mathbf{.783 \pm .026}$ & $\mathbf{.634 \pm .047}$ \\
AttnSleep              & Raw 1D + MRCNN + attention     & $.764 \pm .030$ & $.603 \pm .048$ \\
WIPSNet-CWT ($D{=}4$)  & CWT + 3D CNN                   & $.747 \pm .032$ & $.556 \pm .035$ \\
WIPSNet-CWT ($D{=}1$)  & CWT + 3D CNN                   & $.744 \pm .022$ & $.527 \pm .026$ \\
WIPSNet-CWT ($D{=}32$) & CWT + 3D CNN                   & $.744 \pm .042$ & $.546 \pm .049$ \\
WIPSNet-CWT ($D{=}2$)  & CWT + 3D CNN                   & $.742 \pm .035$ & $.558 \pm .044$ \\
WIPSNet-CWT ($D{=}16$) & CWT + 3D CNN                   & $.738 \pm .036$ & $.535 \pm .039$ \\
WIPSNet-CWT ($D{=}12$) & CWT + 3D CNN                   & $.716 \pm .030$ & $.492 \pm .023$ \\
catch22 + XGBoost      & Raw 1D, 22 canonical TS feats  & $.704 \pm .002$ & $.481 \pm .002$ \\
WIPSNet-CWT ($D{=}24$) & CWT + 3D CNN                   & $.701 \pm .038$ & $.505 \pm .056$ \\
Mamba ($32$\,min)      & Raw 1D + SSM                   & $.689 \pm .047$ & $.509 \pm .060$ \\
Mamba ($4$\,min)       & Raw 1D + SSM                   & $.670 \pm .053$ & $.479 \pm .034$ \\
WIPSNet-WST ($D{=}8$)  & Wavelet scattering + 3D CNN    & $.670 \pm .039$ & $.460 \pm .029$ \\
Mamba ($16$\,min)      & Raw 1D + SSM                   & $.659 \pm .068$ & $.506 \pm .060$ \\
EVI                    & Clinical scalar                & $.633$          & $.433$          \\
Mamba ($8$\,min)       & Raw 1D + SSM                   & $.613 \pm .106$ & $.459 \pm .116$ \\
WIPSNet-STFT ($D{=}8$) & STFT + 3D CNN                  & $.578 \pm .080$ & $.505 \pm .086$ \\
SleePyCo               & Raw 1D + pyramid + Transformer & $.577 \pm .152$ & $.444 \pm .101$ \\
Mamba-Scalogram        & CWT patches + SSM              & $.569 \pm .138$ & $.441 \pm .104$ \\
XGBoost                & Handcrafted CWT (44 feat)      & $.553 \pm .010$ & $.402 \pm .006$ \\
\bottomrule
\end{tabular}
\end{table}